\documentclass[10pt,letterpaper]{article}

\usepackage{cogsci}

\cogscifinalcopy 
\usepackage{graphicx}
\usepackage{amsmath}
\usepackage{pslatex}
\usepackage{apacite}
\usepackage{float} 

\newcommand\blfootnote[1]{%
  \begingroup
  \renewcommand\thefootnote{}\footnote{#1}%
  \addtocounter{footnote}{-1}%
  \endgroup
}

\usepackage[labelfont=bf]{caption}
\usepackage{booktabs}

\title{Belief Attribution as Mental Explanation: \\ The Role of Accuracy, Informativity, and Causality}
 
\author{\\{\large \bf Lance Ying$^{*1,2}$, Almog Hillel$^{*1}$, Ryan Truong$^{*2}$} \\
{\large \bf Vikash K. Mansinghka$^{1}$, Joshua B. Tenenbaum$^{1}$, Tan Zhi-Xuan$^{1}$} \\ \\
  $^{1}$Massachusetts Institute of Technology, Cambridge, MA, USA\\
  $^{2}$Harvard University, Cambridge, MA, USA\\
  $^{*}$Equal Contribution\\ \\
  }
\begin{document}

\maketitle

\begin{abstract}
A key feature of human theory-of-mind is the ability to attribute beliefs to other agents as mentalistic explanations for their behavior. But given the wide variety of beliefs that agents may hold about the world and the rich language we can use to express them, which specific beliefs are people inclined to attribute to others? In this paper, we investigate the hypothesis that people prefer to attribute beliefs that are \emph{good explanations} for the behavior they observe. We develop a computational model that quantifies the explanatory strength of a (natural language) statement about an agent's beliefs via three factors: accuracy, informativity, and causal relevance to actions, each of which can be computed from a probabilistic generative model of belief-driven behavior. Using this model, we study the role of each factor in how people selectively attribute beliefs to other agents. We investigate this via an experiment where participants watch an agent collect keys hidden in boxes in order to reach a goal, then rank a set of statements describing the agent's beliefs about the boxes' contents. We find that accuracy and informativity perform reasonably well at predicting these rankings when combined, but that causal relevance is the single factor that best explains participants' responses.

\textbf{Keywords:} 
theory of mind; mental explanation; belief attribution; causal attribution; communication; social inference

\end{abstract}

\section{Introduction}
Suppose you are watching your friend Sally upend her fruit basket in an effort to find her lost keys, when your mutual friend Anne walks in and asks what in the world Sally is thinking. How would you be inclined to describe Sally's beliefs? Although there are a great many beliefs that Sally likely holds --- that she owns a fruit basket, that her fruit basket is now upended, and that she has lost her keys --- none of these seem quite as apt as the explanation that she believes she might have lost her keys in her fruit basket. \blfootnote{Presented as a conference paper at CogSci 2025}
This tendency to attribute beliefs that are \emph{good explanations} for other agents' behavior is highly intuitive, yet understudied. While a large body of literature has investigated how people infer and attribute beliefs to others by exercising their theory-of-mind \shortcite{wimmer1983beliefs,onishi200515,baker2017rational}, and a similarly large of body of work has studied how people generate and evaluate explanations \shortcite{lombrozo2006structure,lombrozo2007simplicity,kirfel2024explain}, prior research on how people choose between mental explanations has focused on agent's goals, intentions and their responsibility for outcomes \shortcite{lagnado2013causal,alicke2015causal,chandra2024cooperative,wu2023computational}. In contrast, selective belief attribution and its relationship with explanation generation have received limited attention.

\begin{figure}
    \centering
    \includegraphics[width=\linewidth]{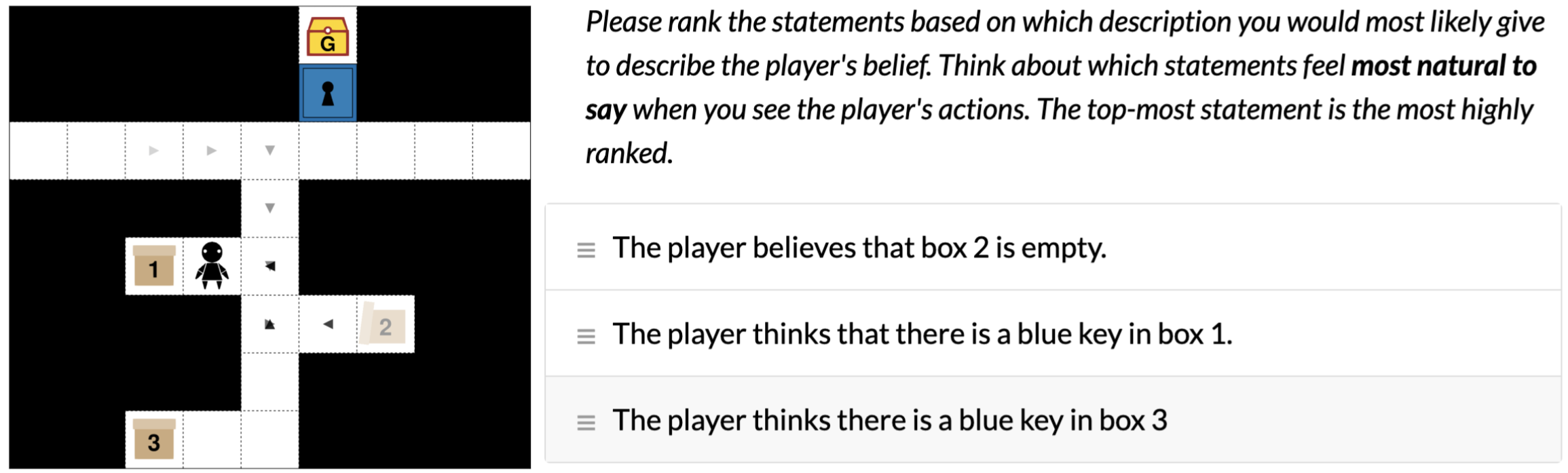}
    \caption{An example scenario in our belief attribution experiment. An animation on the left shows a player in a treasure game, who is trying to find a blue key to unlock the blue door to retrieve the gold chest. On the right, participants are asked to rank three statements about the player's beliefs based on their likelihood of attributing each statement.}
    \vspace{-12pt}
    \label{fig:interface}
\end{figure}

In this paper, we investigate how and whether people attribute beliefs on the basis of their explanatory strength, developing a computational model of selective belief attribution that scores and selects candidate beliefs. We represent candidate belief attributions in natural language, building upon recent extensions of the Bayesian Theory-of-Mind (BToM) framework that provide a grounded compositional semantics for belief statements \shortcite{ying2025understanding,ying2024grounding}. Unlike the non-linguistic belief representations in earlier work, \shortcite{baker2017rational}, this allows us to study how people attribute belief statements that highlight specific aspects of others' minds. To quantify the explanatory strength of a belief statement, we take inspiration from accounts of explanation as probabilistic inference \shortcite{lombrozo2007simplicity,wojtowicz2020probability}, rational communication \shortcite{chandra2024cooperative,kirfel2024explain} and causal judgment \shortcite{gerstenberg2015whether,gerstenberg2020expectations,quillien2023counterfactuals}, factoring explanatory strength into accuracy, informativity, and causal relevance to an agent's actions. These factors are computed as probabilistic or causal quantities from a generative model of rational belief-driven action.

To evaluate our model, we design an experiment where participants are shown animations of a player character navigating a gridworld puzzle, which requires the player to find keys hidden in boxes that unlock doors to their goal (see Figure \ref{fig:interface}). We recruited participants to watch the player's actions, and present them with three statements that describe the player's beliefs about the boxes' contents. Participants are asked to rank these statements based how likely they would attribute each belief, without any prompting to do so on the basis of explanatory strength. Our model also ranks these statements by each of the explanatory factors, and combinations thereof. By comparing these rankings, we find that accuracy and informativity predict human rankings reasonably well when combined, but that causal relevance is the single factor that best explains how people selectively attribute beliefs.

\section{Related Work}

\paragraph{Computational Models of Belief Attribution}

Computational accounts of belief attribution include approaches based on epistemic logic \shortcite{shvo2020epistemic,cedegao2021does} and Bayesian theory-of-mind (BToM) \cite{baker2017rational}. Our model builds on the latter framework, which models how observers infer beliefs as Bayesian inference over a generative model of how goals and beliefs guide actions. We leverage a recent extension of this framework that can infer beliefs represented in \emph{epistemic language} \shortcite{ying2024grounding,ying2025understanding}, allowing us to represent a much richer set of belief attributions. While prior work in the BToM paradigm focuses on assessing the probability that an agent holds a specific belief, we study the \emph{selective attribution} of beliefs, which we model as explanatory choice. To do so, we use the fact that a generative theory-of-mind is also a causal model, allowing us to evaluate the effect of hypothetical interventions on agents' minds \shortcite{ho2022planning,chen2024intervening,wu2024change}. We use this to quantify the causal relevance of a belief on an agent's actions, alongside non-causal quantities.

\paragraph{Generating and Evaluating Explanations}

Our model draws upon several accounts how people evaluate or select between explanations. Bayesian accounts capture how people adjudicate between multiple competing explanations \shortcite{lombrozo2007simplicity,wojtowicz2020probability}. Communicative accounts model whether an explanation is helpful to an (implied) listener \shortcite{chandra2024cooperative,machino2024listener}, pragmatically trading off between accuracy and informativity \shortcite{scontras2021practical,quillien2022logic}. Causal accounts emphasize the fact good explanations are not just probable, but also causally relevant to the observations explained \shortcite{gerstenberg2015whether,gerstenberg2021counterfactual,quillien2023counterfactuals}. Each of these accounts is reflected in the explanatory factors computed by our model, allowing us to study their influence on belief attribution. Recent work has also combined these accounts, modeling explanation as the communication of useful interventions \shortcite{kirfel2022inference,kirfel2024explain}.

\section{Computational Model}

Our model builds on the language-augmented Bayesian theory-of-mind (LaBToM) framework of \citeA{ying2025understanding}, which combines (i) a generative model of how agents rationally take sequences of actions given their beliefs, and (ii) a formal language for representing statements about agent's beliefs. We use these components to compute (iii) the accuracy, informativity, and causal relevance of a belief statement as explanatory factors that might influence an observer's preference for attributing some belief statements over others. Using these factors, we can (iv) (probabilistically) rank belief statements based on their explanatory strength.

\subsection{Modeling Rational Action and Perception}

Our generative model follows the structure of Bayesian theory-of-mind models \shortcite{baker2017rational,ying2023neuro}, specialized to the case where the agent's goal $g$ (i.e. reaching the treasure chest) is known to the observer:
\begin{alignat*}{2}
    \textit{State Prior:}& \quad S_0 \sim P(S_0) \\
    \textit{Belief Prior:}& \quad B_0 \sim P(B_0 | S_0) \\
    \textit{State Transition:} & \quad S_t \sim P(S_t | S_{t-1}, A_t) \\
    \textit{Belief Update:}& \quad B_t \sim P(B_t | S_t, B_{t-1})  \\
    \textit{Action Selection:}& \quad A_t \sim P(A_t | B_t; g) \\
    \textit{Observations:}& \quad O_t \sim P(O_t | S_t)
\end{alignat*}

This model specifies how the agent takes an action $A_t$ at each step $t$ given their beliefs about the environment state $S_t$, while updating their belief $B_t$ based on what they observe. Actions follow a Boltzmann-rational distribution:
\begin{equation*}
    P(A_t | B_t; g) \propto \exp\left(-\hat Q_g(B_t, A_t)\right)
\end{equation*}
where $\hat Q_g(B_t, a)$ is an estimate of the cost of reaching the goal $g$ via action $a$ from belief state $B_t$. As such, if the agent believes that a key necessary to reaching the goal $g$ is hidden in some box, actions leading to that box will be more probable. A belief $B_t$ is represented as probability distribution $\{(\tilde S_i, W_i)\}_{i=1}^{n_s}$ over $n_s$ possible states, where $W_i$ is the probability assigned to state $\tilde S_i$. This belief is updated deterministically based on consistency with what the agent observes.

Finally, our model assumes that the observer does not have full visibility of the environment $S_t$, and instead only observes the agent's action $A_t$ and a partial observation $O_t$ (i.e. everything in $S_t$ except the contents of unopened boxes). We also assume a uniform prior over initial environment states $P(S_0)$ (corresponding to all ways up to 2 keys can be placed in boxes while ensuring the goal is reachable), and an initial belief prior $P(B_0)$ that is uniform over all ways of distributing $K=3$ equal weights over the initial states. More details about this model can be found in \citeA{ying2025understanding}.

\subsection{Representing and Evaluating Belief Statements}

We use the epistemic language-of-thought (ELoT) formalism introduced in \citeA{ying2025understanding} to represent natural language statements about what an agent knows or believes. For example, the statement \emph{``The player knows that the red key is in either box 1 or box 2''} can be translated into the epistemic formula $\varphi := \textsf{knows}(\textsf{player}, \phi) \equiv \textsf{believes}(\textsf{player}, \phi) \land \phi$, where $\phi := \exists K. \textsf{iscolor}(K, \textsf{red}) \land (\textsf{inside}(K, \textsf{box1}) \lor \textsf{inside}(K, \textsf{box2}))$ is an embedded (non-epistemic) formula. The ELoT formalism adopts a probabilistic semantics for modal operators like \textsf{believes}, so $\varphi$ is interpreted as the claim that $\textbf{Pr}(\textsf{player}, \phi) \geq \theta_\textsf{believes} \land \phi$ --- in words, \emph{``The player assigns a probability greater than some threshold $\theta_\textsf{believes}$ to the proposition $\phi$, and $\phi$ is true''}. See \citeA{ying2025understanding} for threshold values, a fuller description of ELoT's compositional semantics, and how natural language can be translated into ELoT formula.

Importantly, an epistemic formula $\varphi$ can be viewed as a Boolean function: Given an agent's belief state $B_t$ and environment state $S_t$, $\varphi(S_t, B_t)$ is true if and only if the formula is true description of the probability distribution represented by $B_t$, and any factive clauses in $\varphi$ are true in $S_t$ (e.g. if $\varphi$ is a knowledge claim). Equivalently, we can define a random variable $\varphi_t := \varphi(S_t, B_t)$ that denotes the truth of $\varphi$ at step $t$. As such, when reasoning about an agent's actions or beliefs with our generative model, we can condition on $\varphi_t$ being false or true to evaluate probabilistic or causal quantities.

\subsection{Explanatory Factors in Belief Attribution}

We can now formalize several factors that might influence belief attribution in terms of probabilistic, information-theoretic, or causal quantities:

\subsubsection{Accuracy as Posterior Probability.}

When attributing a belief statement $\varphi$ to an agent, an obvious candidate for what an observer might prioritize is \emph{accuracy} --- that the statement $\varphi$ is (probably) true based on the available evidence.
With our generative model, we can formalize the accuracy \textsf{Acc} of statement $\varphi$ at step $t$ as the posterior probability of $\varphi_t$ being true (denoted \textsf{T}) given all actions $a_{1:t}$ and observations $o_{1:t}$:
\begin{multline*}
    \textsf{Acc}(\varphi, t) := P(\varphi_t = \textsf{T} | A_{1:t} = a_{1:t}, O_{1:t} = o_{1:t})\\
    = \textstyle\sum_{\varphi(S_t, B_t) = \textsf{T}} P(S_t, B_t | A_{1:t} = a_{1:t}, O_{1:t} = o_{1:t}) 
\end{multline*}
Following prior work, we use a 50-50 prior over whether $\phi_t$ is true when computing accuracy, avoiding counting effects that lead to poorer fits with human responses \cite{ying2024grounding}.

\subsubsection{Informativity as Information Gain.}

If people attribute beliefs in order to explain an agent's behavior to other observers, then another factor they might prioritize is the \emph{informativity} of the statement $\varphi$ to an assumed listener. We formalize this as the information gain (i.e. the Kullback-Leibler divergence) provided by a statement $\varphi$ about the agent's beliefs $B_t$ (and also the environment $S_t$, since knowledge claims can be factive), relative to a listener who can also observe the environment but has not drawn inferences from the agent's actions:\footnote{Since our experiment restricts the statements that participants rank to those about the player's beliefs, it is natural to model the listener as knowing about the environment, but ignorant of the beliefs that can be inferred from prior actions. We also consider an entirely ignorant listener, but this results in a worse fit (Table \ref{tab:corr}).}
\begin{multline*}
    \textsf{Info}(\varphi, t) := \\
    \operatorname{KL}[P(S_t, B_t|\varphi_t = \textsf{T}, O_t = o_t) || P(S_t, B_t| O_t = o_t)]
\end{multline*}
By combining accuracy and informativity, an observer can avoid attributing beliefs that are trivially true, which would be unhelpful to a listener.

\subsubsection{Causal Relevance as Hypothetical Reasoning.}

While accuracy and informativity prioritize beliefs regardless of how they influence agent's actions, \emph{causal relevance} specifically accounts for whether a belief is a cause for such actions. For example, in Figure \ref{fig:interface}, it seems more natural to explain the player's movement to box 1 with the statement \emph{``The player thinks that there is a blue key in box 1''}, whereas \emph{``The player believes that box 2 is empty''} seems like a poorer explanation even though it is almost certainly true (and potentially quite informative, depending on what the listener knows). One analysis of this intuition is that believing that a blue key is in box 1 is both a \emph{necessary} and \emph{sufficient} cause for the player to walk towards that box. In contrast, believing that box 2 is empty is at best a necessary cause for those actions, and one that is guaranteed to occur as long as the player is rational.

To formalize this causal intuition, we quantify the \emph{causal necessity} of a belief statement $\varphi$ as the probability that the agent would \emph{not} have taken the observed actions under a hypothetical intervention that sets $\varphi$ to be false.\footnote{For simplicity, we consider hypothetical rather than counterfactual interventions, since they do not come apart significantly in our experiment.} Since $\varphi$ describes the agent's current beliefs, we perform the intervention at the most recent step $t_c$ where the agent's beliefs changed (i.e. when they last opened a box), and evaluate only the subsequent actions while conditioning on the past:
\begin{equation*}
    \textsf{CNecc}(\varphi, t) := P(A_{t_c:t} \neq a_{t_c:t} | \textsf{do}(\varphi_{t_c} = \textsf{F}), a_{1:t_c-1}, o_{1:t_c-1})
\end{equation*}
Analogously, we quantify the \emph{causal sufficiency} of $\varphi$ as the probability that an agent \emph{does} take the actions after $t_c$ when $\varphi$ is set to true at $t_c$:
\begin{equation*}
    \textsf{CSuff}(\varphi, t) := P(A_{t_c:t} = a_{t_c:t} | \textsf{do}(\varphi_{t_c} = \textsf{T}), a_{1:t_c-1}, o_{1:t_c-1})
\end{equation*}
Finally, we follow prior work showing that the \emph{normality} of a cause influences causal judgment, with people preferring to cite abnormal (e.g. less frequent) causes when multiple causes are necessary, and normal causes when multiple are sufficient \cite{icard2017normality,gerstenberg2020expectations}. We quantify normality as the probability of $\varphi_{t_c}$ being true, conditioned on past observations:
\begin{equation*}
    \textsf{CNorm}(\varphi) := P(\varphi_{t_c} = \textsf{T} | a_{1:t_c-1}, o_{1:t_c-1})
\end{equation*}
Following the measure of actual causal strength introduced in \citeA{icard2016causality} and \citeA{icard2017normality}, we combine causal normality, necessity and sufficiency into an overall measure of causal relevance, multiplying the necessity of a belief by its abnormality and sufficiency by its normality:
\begin{align*}
    \textsf{Causal}(\varphi, t) := &[(1-\textsf{CNorm}(\varphi)) \textsf{CNecc}(\varphi, t)]^{\alpha_\textsf{CNecc}} \cdot \\
    &[\textsf{CNorm}(\varphi) \textsf{CSuff}(\varphi, t)]^{\alpha_\textsf{CSuff}}
\end{align*}
This formulation effectively rules out guaranteed beliefs as causes (e.g. believing that a box is empty after seeing that it is empty), even if they are necessary to explain certain actions.

\subsection{Probabilistic Attribution of Belief Statements}

Each of the factors above can combined into an overall score measuring the explanatory strength of a belief statement $\varphi$. For a set of factors $F$, we do this via linear combination of their logarithms (except for \textsf{Info}, which is in log-space):
\begin{equation*}
    \textsf{Score}(\varphi, t, F) := \textstyle\sum_{f \in F} \alpha_f \log(f(\varphi, t))
\end{equation*}
We then model the attribution a belief statement $\varphi$ out of a set of candidate statements $\Phi$ via a probabilistic choice rule:
\begin{equation*}
    P_\text{attribute}(\varphi; \Phi, t) = \exp(\textsf{Score}(\varphi, t)) / \textstyle\sum_{\phi \in \Phi} \textsf{Score}(\phi, t)
\end{equation*}
This attribution model can be used to compute the probability of a ranking $\varphi_1 \succ \varphi_2 \succ ... \succ \varphi_n$ over all statements $\varphi_i \in \Phi$:
\begin{equation*}
    P_\text{rank}(\varphi_1 \succ ... \succ \varphi_n) = 
    \textstyle\prod_{i=1}^n P_\text{attribute}(\varphi_i; \Phi \setminus \phi_{1:(i-1)}, t)
\end{equation*}
This allows us to compute the average rank of each statement as the weighted sum of its position among all rankings. 

\begin{figure*}[ht!]
    \centering
    \includegraphics[width=\linewidth]{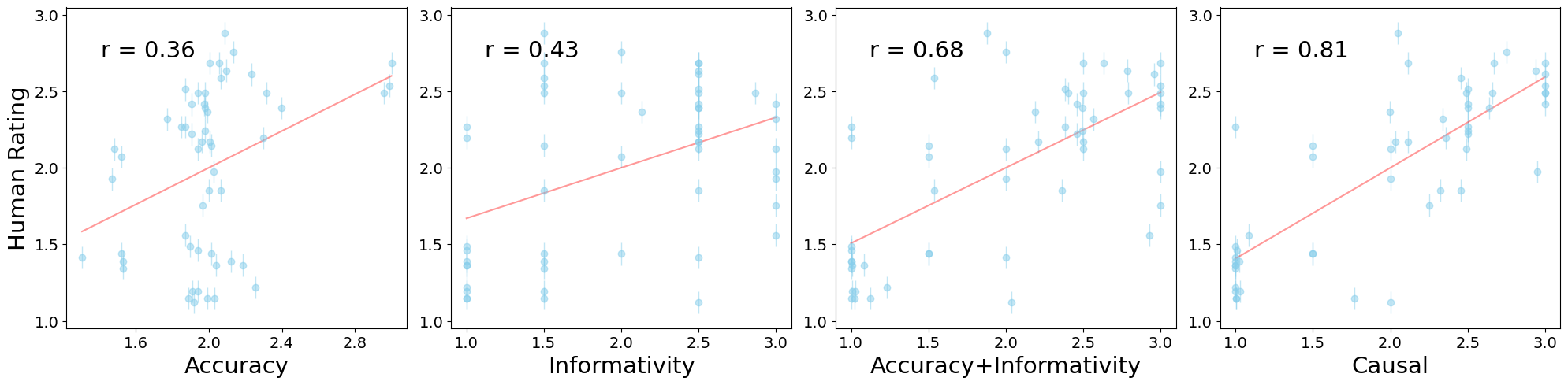}
    \caption{Correlation plots between average human rankings ($y$-axis) and model rankings ($x$-axis) for selected combinations of explanatory factors. Among the factors shown, the Causal model best explains human judgments ($r=0.81$), with Accuracy + Informativity coming second ($r=0.68$).}
    \label{fig:corr_plot}
\end{figure*}

\begin{figure*}[ht!]
    \centering
    \includegraphics[width=\linewidth]{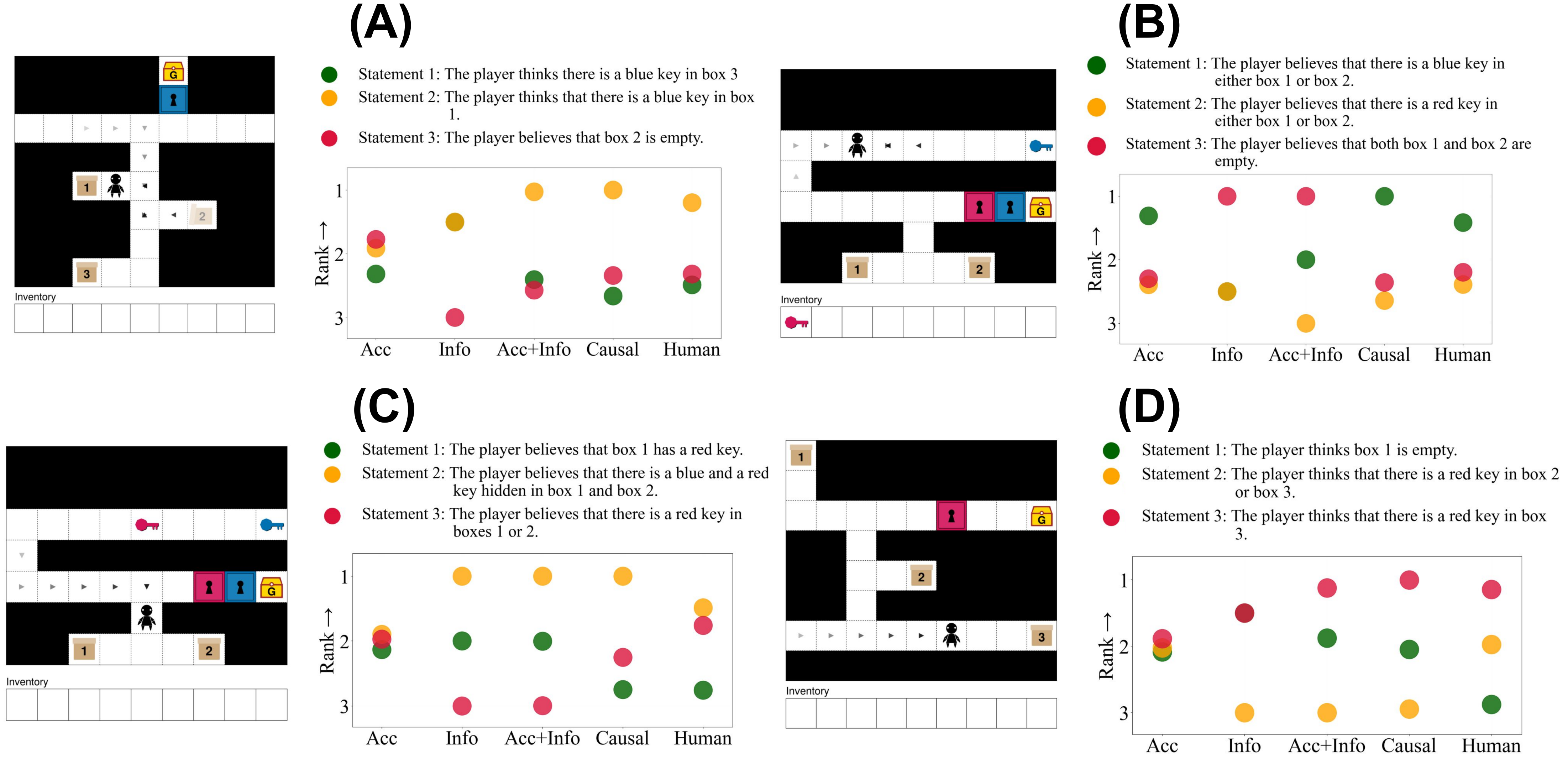}
    \caption{Qualitative examples showing model and human rankings of sets of 3 belief statements. On the left of each example is the agent's trajectory in the enivonment. On the right, we plot a continuous rank metric for each statement and model.}
    \label{fig:qualitative}
    \vspace{-9pt}
\end{figure*}

\section{Experiment}

To evaluate our model of how humans interpret belief statements, we designed an experiment where participants had to infer the beliefs of an agent solving a gridworld puzzle used previously by \citeA{ying2024grounding}. In these puzzles, an agent has to pick up keys and unlock doors to reach a valuable chest. Doors can only be opened by keys of the same color and each key can be used once. Some keys are hidden inside boxes and the player cannot see what's inside the box.

We designed 6 maps, and 3 different key allocations and agent trajectories per map, for a total of 18 scenarios. The scenarios were designed with varying agent beliefs, many of which involves the agent having false beliefs and replanning with new observations, such as opening an empty box.

For each scenario, we designed 3 statements describing the player's beliefs. The statements vary in their content and structure, such as compositionality (e.g. \emph{``The player believes that the red key is in box 1 and a blue key in box 2''}), specificity (e.g. \emph{``The player believes that there must be a red key in box 1''}), and the use of epistemic verbs (e.g. \emph{``The player \textbf{knows} that the red key is in box 1''}). To disambiguate our hypothesized explanatory factors, we carefully chose statements that were all plausible, since we anticipated that humans would largely prefer truthful statements over false ones.

\begin{table}[ht!]
\footnotesize
\centering
\begin{tabular}{@{}rlrrrr@{}}
\toprule
\textbf{Factors}              & Cor. {\scriptsize(95\% CI)}                                              & \multicolumn{1}{l}{$\alpha_\textsf{Acc}$} & \multicolumn{1}{l}{$\alpha_\textsf{Info}$} & \multicolumn{1}{l}{$\alpha_\textsf{CNecc}$} & \multicolumn{1}{l}{$\alpha_\textsf{CSucc}$} \\ \midrule
\textsf{Acc}                  & \begin{tabular}[c]{@{}l@{}}0.36\\ {\scriptsize(0.29--0.42)}\end{tabular} & 1.342                                     & --                                         & --                                          & --                                          \\
\textsf{Info}                 & \begin{tabular}[c]{@{}l@{}}0.43\\ {\scriptsize(0.36--0.48)}\end{tabular} & --                                        & 7.720                                      & --                                          & --                                          \\
\textsf{Causal}               & \begin{tabular}[c]{@{}l@{}}0.81\\ {\scriptsize(0.76--0.82)}\end{tabular} & --                                        & --                                         & 1.600                                       & 10.00                                       \\
\textsf{Acc}+\textsf{Info}    & \begin{tabular}[c]{@{}l@{}}0.68\\ {\scriptsize(0.62--0.70)}\end{tabular} & 10.00                                     & 6.000                                      & --                                          & --                                          \\

\textsf{Acc}+\textsf{Causal}  & \begin{tabular}[c]{@{}l@{}}0.82\\ {\scriptsize(0.77--0.84)}\end{tabular} & 10.00                                     & --                                         & 1.528                                       & 3.056                                       \\
\textsf{Info}+\textsf{Causal} & \begin{tabular}[c]{@{}l@{}}0.81\\ {\scriptsize(0.76--0.82)}\end{tabular} & --                                    & 0.240                                          & 1.000                                       & 10.00                                      \\
\textsf{All Factors}          & \begin{tabular}[c]{@{}l@{}}0.82\\ {\scriptsize(0.77--0.84)}\end{tabular} & 10.00                                     & 0.086                                      & 1.216                                       & 2.432                                       \\ \midrule
\textsf{Info}*                & \begin{tabular}[c]{@{}l@{}}0.35\\ {\scriptsize(0.31--0.39)}\end{tabular} & --                                        & 4.400                                      & --                                          & --                                          \\
\textsf{Acc}+\textsf{Info}*   & \begin{tabular}[c]{@{}l@{}}0.54\\ {\scriptsize(0.49--0.57)}\end{tabular} & 10.00                                     & 7.000                                      & --                                          & --                                          \\ \bottomrule
\end{tabular}
\caption{Correlations between human and model rankings of belief statements for each combination of factors, alongside the fitted parameters $\alpha_f$. \textsf{Info}* is an alternative informativity measure for listeners with no visibility of the environment.}
\vspace{-9pt}
\label{tab:corr}
\end{table}

\subsection{Experiment Design}
Our experiment was conducted via a customized web interface (Figure \ref{fig:interface}). Participants went through a tutorial and comprehension quiz before proceeding to the main experiment. Each participant completed all 21 scenarios in a randomized order. In each scenario, the participant first watched the agent take actions in the map, then was asked to ranked three belief statements based on their likelihood of attributing each statement to the agent. Specifically, we provide participants with the following prompt, which avoids framing the task in terms of explanation or communication to a specific listener: \emph{``Please rank the statements based on which description you would most likely give to describe the player's belief. Think about which statements feel \textbf{most natural to say} when you see the player's actions.''}

\subsection{Participants}
We recruited 41 US participants through Prolific (mean age = 39.39, 14 female, 25 male, 2 non-binary). The experiment took 17 minutes on average.

\subsection{Model Fitting}

To evaluate the fit of our model with participants' responses, we first computed the averaged human ranking for each statement by taking the average of the statement's rank among all participants. As described in the modeling section, we compute the average rank assigned by the model by first computing the probability of all 6 possible rankings of the three statements, then taking the probability-weighted sum over these rankings. We then fit the coefficients $\alpha_f$ for each explanatory factor by maximizing the Pearson's correlation between the average predicted ranks and the average human-provided ranks for each statement. Fitting was performed for every combination of the explanatory factors via grid search over the range $[0, 10]$ for each coefficient, and the fitted values are reported in Table \ref{tab:corr}. (Note that the coefficient for the \textsf{Causal} factor is subsumed under the coefficients $\alpha_\textsf{CNecc}$ and $\alpha_\textsf{CSucc}$).

\subsection{Results}

\subsubsection{Correlation Analysis}
The correlations between model and human averaged rankings are reported in Table \ref{tab:corr} for all combinations of explanatory factors, and we plot the results for a subset of these factors in Figure \ref{fig:corr_plot}. For comparison, the split-half correlation among human participants was $r=0.94$. Our results indicate that the causal relevance factor (\textsf{Causal}) was the single factor that best explained human rankings over belief attributions, with a correlation of $r=0.81$. In contrast, neither accuracy (\textsf{Acc}, $r=0.36$) nor informativity (\textsf{Info}, $r=0.43$) on their own provided good predictions of human ratings. However, the combination of accuracy and informativity (\textsf{Acc}+\textsf{Info}) also produced a reasonable fit to the human data ($r=0.68$). Other combinations of factors with the \textsf{Causal} factor only resulted in marginal improvements to the correlation $r=0.82$, albeit with the \textsf{Acc} factor playing a larger predictive role (as measured by the magnitude of $\alpha_\textsf{Acc}$) in the \textsf{Acc}+\textsf{Causal} and \textsf{All Factors} combinations (Table \ref{tab:corr}).

By inspecting the correlation plots (Figure \ref{fig:corr_plot}), we find that both the \textsf{Acc}+\textsf{Info} model and the \textsf{Causal} model can identify the most highly ranked statement that humans did in most scenarios (see clusters at bottom left of each plot). However, ranking the other two statements appears to be more challenging, and both models and humans show significant uncertainty. We also found a high correlation of $r=0.81$ between the \textsf{Acc}+\textsf{Info} model and the \textsf{Causal} model, indicating that many of the explanatory preferences captured by causal relevance are also be captured by maxims of cooperative communication. We speculate that this is because a good causal explanation also tends to be accurate and informative. However, an accurate and informative belief statement may not always be causally relevant. This could explain why the \textsf{Causal} and \textsf{Acc}+\textsf{Info} models performed similarly on many scenarios, yet the \textsf{Causal} model correlated better as a whole.

The performance of accuracy and informativity also hinges on how exactly informativity is modeled: if we instead assume the lister can see nothing about the environment, the resulting informativity metric (\textsf{Info}*, $r=0.35$) and its combination with accuracy (\textsf{Acc}+\textsf{Info}*, $r=0.54$) have substantially lower correlations with human rankings (Table \ref{tab:corr}).

\subsubsection{Qualitative Analysis}

In Figure \ref{fig:qualitative}, we show four qualitative examples comparing model and human rankings. In scenario A, the \textsf{Acc} model ranked statement 3 first, the \textsf{Info} model ranked both statement 1 and 2 to be first, whereas the \textsf{Acc}+\textsf{Info} model and the \textsf{Causal} model ranked statement 2 first, as did human participants. However, the latter two models produce these rankings for different reasons. The \textsf{Acc}+\textsf{Info} model ranks statement 2 first because it's informative and likely to be true. On the other hand, the \textsf{Causal} model produces this ranking because statement 2 is both causally necessary and sufficient to explain why the agent to walked towards box 1 and not box 3. 

Scenario B and C show two cases that highlight the differences in ranking between \textsf{Acc}+\textsf{Info} and the \textsf{Causal} model. In scenario B, the player picks up a red key and walks away from the blue key. The \textsf{Acc}+\textsf{Info} model ranked statement 3 higher than statement 1, whereas the \textsf{Causal} model and humans ranked them in an opposite way. Here the \textsf{Acc}+\textsf{Info} preferred statement 3 as it's more informative than the rest, although it's less accurate because the agent needs a blue key to reach the chest and the player is likely looking for it among the two boxes. In scenario C, the \textsf{Acc}+\textsf{Info} model ranked statement 1 higher than statement 3, whereas the \textsf{Causal} model and humans ranked in the other direction. This is because statement 1 is more informative than statement 3, even though statement 3 has higher accuracy than statement 1. These two scenarios show that the \textsf{Acc}+\textsf{Info} model may have difficulty in deciding between high accuracy and low informativity versus low accuracy and high informativity statements.

Lastly, we show an error case in scenario D. In this scenario, both \textsf{Acc}+sf{Info} and \textsf{Causal} models fail to match how humans rank statements 1 and 2.  The \textsf{Acc}+\textsf{Info} model ranks statement 2 last because it is less informative than statement 1. The \textsf{Causal} model assigns statement 2 a low score because if the statement were true (i.e. if the player assigned a high enough probability to the red key being in \emph{either} box 2 or 3, even if they were uncertain as to which), a rational player would first go to check box 2, since box 2 is closer to both the player and the goal. This results in low likelihood of the observed actions under the intervention that statement 2 is true. This suggests that our model may need to better account boundedly rational behavior (e.g. the agent pursuing the sub-optimal sub-goal of checking box 3), or alternative readings of disjunctive statements (statement 2 could also be read as stating that the player thinks that a red key is in box 2, \emph{or} that the player thinks the key is in box 3).

\section{Discussion}

In this paper, we examined how people selectively attribute beliefs in light of others' behavior. While prior work has studied both belief attribution and explanation generation separately, our findings bridge these domains by demonstrating that people's belief attributions are guided by explanatory considerations. In particular, we found that causal relevance was the single strongest predictor of which beliefs people attribute to others, although a combination of accuracy and informativity also provided reasonable predictions of human judgments. Nonetheless, the fact that causal relevance provided strong predictions suggests that people may naturally gravitate toward beliefs that causally explain actions, especially in cases without an explicit communicative context. This interpretation is further supported by our finding that when we assume a listener with no visibility of the environment, the predictive power of accuracy and informativity decreases substantially.

However, several important limitations and open questions remain. First, while our stimuli allowed us to demonstrate the importance of explanatory factors beyond accuracy and informativity, they do not strongly disambiguate between causal relevance and the combined effects of accuracy and informativity in many scenarios. Future work could develop more targeted paradigms that tease apart these factors.

Second, there are many ways of formalizing causal relevance \shortcite{tian2000probabilities,pearl2009causality,gerstenberg2015whether,icard2017normality,quillien2023counterfactuals}, and it remains unclear which formulation best captures people's casual intuitions about mental states. More careful experimental design might allow us to distinguish a hypothetical vs. counterfactual formulation of causal relevance \shortcite{gerstenberg2021counterfactual}, or a necessity-sufficiency model \shortcite{icard2017normality} vs. a counterfactual effect-size model \shortcite{quillien2023counterfactuals}.

Third, the model used in our experiment assumes that (i) only one possible reading of each belief statement is valid (even though disjunctive statements often have multiple readings), (ii) the agent is Boltzmann rational, which does not account for sub-optimal sub-goal selection (e.g. the possibility of looking in a box that is further away to find a key, before looking in a closer box, as in Figure \ref{fig:qualitative}D). Future work could integrate both ambiguity-aware parsing \shortcite{saparina2024ambrosia} of epistemic statements and boundedly-rational planning \shortcite{zhi2020online,alanqary2021modeling}, potentially improving the model's fit to human judgments.

Fourth, our results shows the potential value of unifying the causal and communicative aspects of explanation selection. Recent work by \citeA{kirfel2024explain} provides one promising direction, suggesting that people select explanations that are causally relevant to the listeners. Extending this approach to mental state attribution would require consideration of what makes belief explanations practically useful, which can depend on the listener's knowledge and goals \shortcite{chandra2024cooperative,machino2024listener}, along with individual factors and the broader social context \shortcite{ying2025benchmarking}.

Lastly, our findings highlight just how much remains unknown about the principles governing which mental explanations we prefer to attribute. Future research could investigate whether these preferences vary across types of mental states (e.g., beliefs vs. desires) or different social contexts (e.g. contexts where agents have to justify or critique others' actions). Understanding these patterns could provide valuable insights into the structure of everyday psychological explanation.

\section*{Acknowledgments}

This work was funded in part by Schmidt AI 2050, the DARPA Machine Common Sense, AFOSR, and ONR Science of AI programs, along with the MIT-IBM Watson AI Lab, the Siegel Family Foundation, and an anonymous donor. Tan Zhi-Xuan is supported by an Open Philanthropy AI Fellowship.

\bibliographystyle{apacite}

\setlength{\bibleftmargin}{.125in}
\setlength{\bibindent}{-\bibleftmargin}

\bibliography{paper}

\end{document}